%% file: neurips_2026.tex
\pdfoutput=1
\documentclass{article}

\PassOptionsToPackage{numbers, compress}{natbib}

 \usepackage[main, final]{neurips_2026}

\usepackage[utf8]{inputenc}
\usepackage[T1]{fontenc}
\usepackage{hyperref}
\usepackage{url}
\usepackage{booktabs}
\usepackage{amsfonts}
\usepackage{nicefrac}
\usepackage{microtype}

\usepackage{graphicx}
\usepackage{amsmath}
\usepackage{amssymb}
\usepackage{multirow}
\usepackage{caption}
\usepackage[table]{xcolor}
\usepackage{bm}
\usepackage{xspace}
\newcommand{\eg}{e.g.\xspace}

\newcommand{\etal}{\textit{et al.}\xspace}

\title{RacerGS: Residual-Autoregressive Context for 3D Gaussian Splatting Compression}

\author{%
  Wenqing Wang \qquad Huimin Zeng \qquad Yun Fu \\
  Northeastern University \\
  \texttt{\{wang.wenqin, zeng.huim, raymondfu\}@northeastern.edu} \\
}

\begin{document}

\maketitle

\begin{center}
  \includegraphics[width=\linewidth, trim=0 85 0 66, clip]{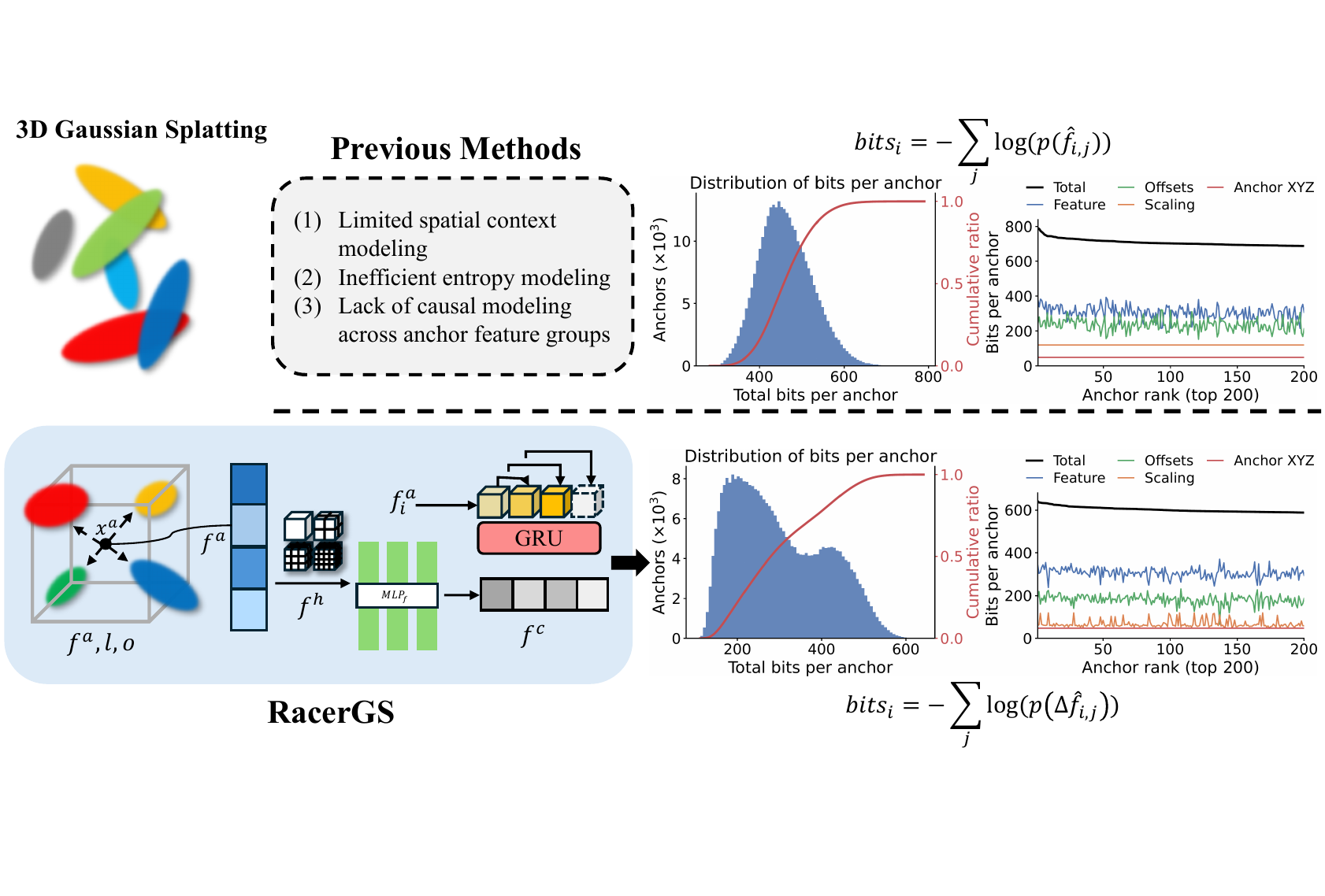}
\captionof{figure}{\textbf{RacerGS for efficient 3DGS compression.}
Previous methods have limited spatial context, inefficient entropy modeling, and lack of causal dependencies across anchor feature groups, leading to high coding costs. RacerGS addresses these limitations through compact context projection, residual coding, and autoregressive GRU refinement. This shifts the bitrate distribution toward fewer bits per anchor and lowers total coding costs while preserving rendering fidelity.}
  \label{teaser}
\end{center}

\input{sec/0_abstract}
\input{sec/1_intro}
\input{sec/2_related_work}
\input{sec/3_method}
\input{sec/4_experiments}
\input{sec/5_conclusion}

\bibliography{main}
\bibliographystyle{plainnat}

\clearpage
\appendix
\input{sec/X_suppl}

\end{document}

%% file: sec/0_abstract.tex
\begin{abstract}
3D Gaussian Splatting (3DGS) has emerged as a promising approach for fast, high-quality novel-view rendering. However, its large parameter size remains a bottleneck for storage and transmission. We propose \textit{RacerGS}, a compact 3DGS compression framework that achieves scene reconstruction at significantly reduced storage sizes with preserved fidelity. To obtain compact and spatially coherent features, we design a hash projection model that projects the anchor hash encodings into compact context features. Additionally, we introduce a FiLM-modulated autoregressive GRU context model that utilizes causal dependencies among anchor feature groups and uses recurrent hidden states to refine entropy-model parameters, improving bitrate efficiency. Furthermore, we adaptively entropy-code quantized anchor-attribute residuals relative to their context-predicted means, leveraging spatial consistency to reduce symbol entropy. Overall, \textit{RacerGS} achieves more than \textbf{$\mathbf{150}\boldsymbol{\times}$} compression over vanilla 3DGS and \textbf{$\mathbf{28}\boldsymbol{\times}$} over Scaffold-GS, while providing comparable rendering quality. Extensive experiments across five datasets demonstrate that \textit{RacerGS} outperforms prior 3DGS compression methods in both storage reduction and rate-distortion performance.
\end{abstract}

%% file: sec/1_intro.tex
\section{Introduction}
\label{sec:intro}

Novel view synthesis has advanced rapidly. Neural Radiance Fields (NeRF)~\cite{NeRF} enable high-quality rendering by modeling scenes as continuous radiance fields, but NeRF-based methods~\cite{NeRF, INGP, TensoRF, K-planes, emogene, realtalk} incur high rendering costs due to dense sampling and repeated field evaluations along each ray.
3D Gaussian Splatting (3DGS)~\cite{3DGS} provides an efficient alternative by explicitly representing scenes with learnable 3D Gaussians and rendering them through tile-based rasterization. This approach enables substantially faster rendering while maintaining high visual quality. However, large-scale scenes can require millions of Gaussians, creating substantial storage and transmission demands.

To compress these large, unstructured Gaussian sets, value-centric methods~\cite{Lee, Simon, EAGLES, Lightgaussian, RDOGaussian, navaneet2024compgs, lee2025compression, liu2024hemgs, xie2025sizegs} reduce parameter count and bitwidth via pruning, quantization, and codebooks, but ignore spatial redundancy. Structure-aware methods~\cite{SOG, mini, SUNDAE, IGS, Contextgs, zhan2025cat3dgs, zhan2025cat3dgspro} instead predict attributes from local neighborhoods. In particular, Scaffold-GS~\cite{scaffold} groups Gaussians around learned anchors and predicts their attributes from anchor features, improving rate--distortion (RD) trade-offs.
Despite these gains, Scaffold-GS models anchors independently, producing sparse, point-cloud-like representations that remain difficult to compress. HAC methods~\cite{HAC, chen2025hac++} capture spatial relations via hash grids, but feed concatenated features directly to entropy networks, increasing cost with dimensionality without explicitly modeling interactions across dimensions. They also quantize anchor attributes directly, ignoring the spatial regularity of residuals relative to predicted means and inflating storage. While HAC++~\cite{chen2025hac++} builds intra-anchor context by repeatedly concatenating decoded chunks, it neither forms a compact history nor captures higher-order dependencies.

We empirically observe that prior methods consume a large number of bits per anchor (Fig.~\ref{teaser}), suggesting room to improve entropy estimation and bit allocation. These observations highlight three challenges: 1) limited spatial context modeling, where hash or anchor features are used directly without capturing structured dependencies; 2) inefficient entropy modeling that ignores the stronger regularity in attribute residuals; and 3) the lack of causal modeling across anchor feature groups, preventing the entropy model from exploiting intra-anchor dependencies.

To address these challenges, we propose \emph{\textbf{RacerGS}}, a unified 3DGS compression framework that jointly leverages structured spatial context, residual entropy modeling, and recurrent causal context refinement. A \textbf{hash projection model} projects anchor hash encodings into compact, fused context features for a stronger prior. A \textbf{FiLM-modulated autoregressive GRU context model} refines entropy parameters via recurrent hidden states over causally ordered feature groups. Finally, \textbf{adaptive residual quantization} encodes residuals around their context-predicted means, yielding lower-entropy symbols and improved RD performance.

Our contributions are summarized as follows:
\begin{itemize}
  \item We propose \emph{RacerGS}, a 3DGS compression framework that achieves highly compact representations while preserving high-fidelity rendering quality.

  \item We address existing 3DGS compression challenges with a hash projection model, a FiLM-modulated autoregressive GRU context model, and adaptive residual quantization for entropy coding, which jointly enhance context modeling and tighten entropy estimation.

  \item Extensive experiments on diverse datasets demonstrate that our method achieves superior 3DGS compression (\eg, \textbf{$\mathbf{150}\boldsymbol{\times}$} over 3DGS; \textbf{$\mathbf{28}\boldsymbol{\times}$} over Scaffold-GS) while preserving rendering quality (\eg, a \textbf{$\mathbf{2.31}$ dB} PSNR gain over 3DGS; \textbf{$\mathbf{0.56}$ dB} over Scaffold-GS).
\end{itemize}

%% file: sec/2_related_work.tex
\section{Related Work}
\label{sec:related_work}

\noindent\textbf{Neural radiance fields and compression.}
Neural Radiance Fields~\cite{NeRF} achieve high-quality novel view synthesis but rely on dense sampling and large MLPs that limit real-time rendering. Explicit representations such as K-planes~\cite{K-planes}, Instant-NGP~\cite{INGP}, and TensoRF~\cite{TensoRF} replace much of the MLP with structured grids, trading compute for storage.
NeRF compression broadly splits into \emph{value-based} approaches (pruning~\cite{Re:NeRF, VQRF}, vector quantization~\cite{BiRF}, codebooks~\cite{CompactNeRF}, and entropy-aware training~\cite{SHACIRA}) and \emph{structure-based} approaches that exploit spatial correlations via factorization~\cite{CCNeRF, MaskDWT}, predictive coding~\cite{SPC-NeRF}, or transform coding~\cite{cnc2024, NeRFCodec, codecnerf}. Value-based approaches primarily reduce the precision or redundancy of stored parameters, while structure-based approaches expose dependencies through the organization of scene information. Both seek to balance representation size against reconstruction quality, with compression efficiency depending on how effectively the representation captures and exploits redundancy. These principles motivate compression methods for explicit 3D scene representations.

\noindent\textbf{3D Gaussian splatting and compression.}
3D Gaussian Splatting (3DGS)~\cite{3DGS} enables fast, high-fidelity rendering with learnable Gaussians, but its large parameter count leads to high storage costs. Existing compression methods can be broadly divided into value-based~\cite{Lightgaussian, Navaneet, lee2025compression, Simon, navaneet2024compgs, liu2024hemgs, xie2025sizegs} and structure-based approaches~\cite{SOG, SUNDAE, mini, IGS, scaffold, Contextgs, chen2025hac++, liu2024compgs, zhan2025cat3dgs, zhan2025cat3dgspro}. Value-based methods reduce attribute redundancy via codebooks and vector quantization~\cite{navaneet2024compgs}, entropy modeling~\cite{liu2024hemgs}, or feature-plane projections for standard codecs~\cite{lee2025compression}. Structure-aware methods exploit spatial organization, such as multi-level grids~\cite{IGS}, anchor-based prediction~\cite{scaffold}, anchor context modeling~\cite{Contextgs}, and hash-grid context for anchor quantization~\cite{HAC, chen2025hac++}. However, HAC-style methods rely on poorly scaled hash contexts, miss higher-order causal dependencies, and encode raw attributes instead of spatially regular residuals, increasing entropy and bitrate. We address these limitations with structured spatial context (a learned projection rather than raw hash features), residual entropy modeling (prediction-centered residuals rather than attributes), and recurrent causal context refinement (a constant-cost recurrent state rather than growing concatenation).

%% file: sec/3_method.tex
\section{Preliminaries}

\noindent\textbf{3D Gaussian Splatting} \cite{3DGS} models a scene with a collection of 3D Gaussians characterized by opacity $\alpha$, color $c$, a 3D covariance matrix $\bm{\Sigma}\in\mathbb{R}^{3\times3}$, and a center position $\bm{\mu}\in\mathbb{R}^{3}$:
\begin{equation}
    G(\bm{x}) = e^{-\frac{1}{2}(\bm{x} - \bm{\mu})^\top \bm{\Sigma}^{-1}(\bm{x} - \bm{\mu})},
\end{equation}
where $\bm{x}\in\mathbb{R}^{3}$ denotes the 3D point position, and $\bm{\Sigma}$ is defined as $\bm{R}\bm{S}\bm{S}^{\top}\bm{R}^{\top}$ with the rotation matrix $\bm{R}\in\mathbb{R}^{3\times3}$ and scaling matrix $\bm{S}\in\mathbb{R}^{3\times3}$. The pixel color $C$ is computed by alpha compositing: $C = \sum_{i \in N} c_i \alpha_i \prod_{j=1}^{i-1} (1 - \alpha_j)$.

\noindent\textbf{Scaffold-GS} \cite{scaffold} introduces an anchor-based 3DGS approach that optimizes the storage requirements and preserves fidelity. It utilizes anchors to group 3D Gaussians and predict their attributes from their associated anchor attributes with MLPs. Each anchor is defined by a position $\bm{x}^a\in\mathbb{R}^3$ and attributes $\mathcal{A}=\{\bm{f}^a\in\mathbb{R}^{D^a}, \bm{l}\in\mathbb{R}^6, \bm{o}\in\mathbb{R}^{3K}\}$, corresponding to anchor features, scaling parameters, and offset parameters, respectively. During rendering, $\bm{f}^a$ is fed to MLPs to predict the attributes of the $K$ Gaussians, whose positions are $\bm{x}^a$ plus the scaled offsets $\bm{o}$, and $\bm{l}$ regulates their positions and shapes.

\section{Method}
\label{sec:method}

\begin{figure}[t!]
  \centering
  \includegraphics[width=\linewidth, trim=0 0 0 0, clip]{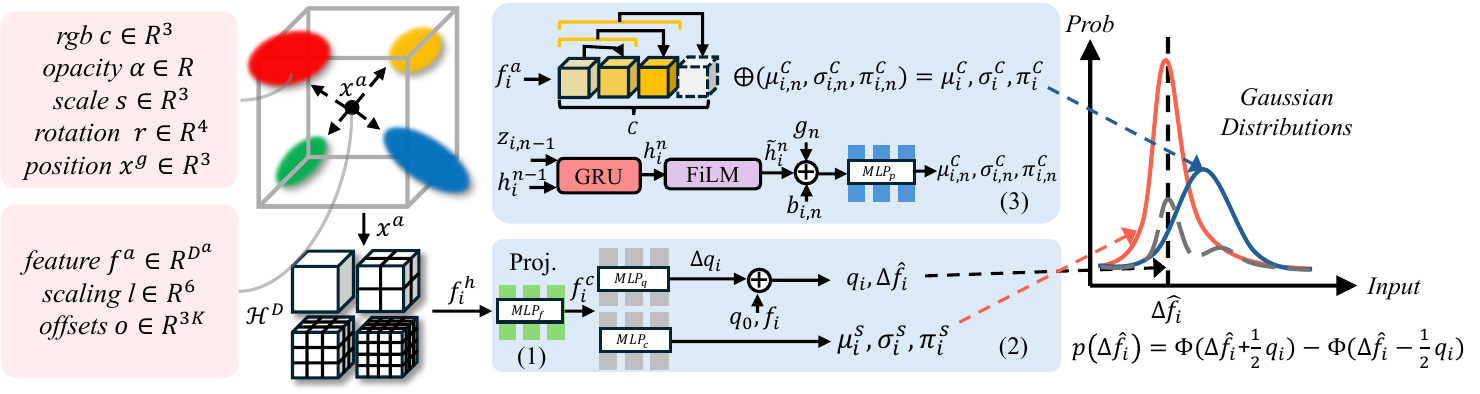}
\caption{\textbf{Pipeline overview.} \textit{RacerGS} compresses 3DGS through three components: \textbf{(1)} a hash projection model maps hash encodings $\bm{f}^h_i$ to compact spatial context features $\bm{f}^c_i$; \textbf{(2)} adaptive residual quantization encodes anchor attributes relative to context-predicted means using learned quantization steps; \textbf{(3)} an autoregressive GRU summarizes previously decoded feature chunks, followed by FiLM modulation and a shared predictor conditioned on base statistics and chunk embeddings. The refined parameters are combined with the spatial prior to form an accurate Gaussian mixture model for residual entropy coding.}
  \label{pipeline}
\end{figure}

\subsection{Overview}
\label{subsec:overview}

As illustrated in Fig.~\ref{pipeline}, \textit{RacerGS} compresses 3DGS by projecting anchor hash encodings into compact context features and predicting entropy model parameters for anchor attribute residuals. For each anchor, a hash projection model extracts context features, while a FiLM-modulated autoregressive GRU recurrently refines per-group distribution parameters using previously decoded feature groups. RacerGS then entropy-encodes quantized residuals around their context-predicted means, reducing coding entropy.

\subsection{Context feature projection}

Using the raw hash features $\bm{f}^h$ directly as the entropy context~\cite{HAC, chen2025hac++} ties the input width of every entropy MLP to the hash dimensionality and forces one encoding to serve both rendering and entropy modeling, whose capacity demands conflict. To obtain spatially coherent context features at the anchor position $\bm{x}^a\!\in\!\mathbb{R}^3$, we instead first query the multi-dimensional hash grid $\mathcal{H}$ to obtain the anchor hash encodings $\bm{f}^h_i=\mathrm{Interp}(\bm{x}^a_i, \mathcal{H})\in\mathbb{R}^{D^h}$. Then, we project $\bm{f}^h_i$ with the hash projection model $\mathrm{MLP_f}$ into the compact fused context features:
\begin{equation}
\bm{f}^{c}_i = \mathrm{MLP_f}(\bm{f}^h_i)\in\mathbb{R}^{D^c}.
\end{equation}
$\bm{f}^c_i$ is the only signal the downstream entropy model receives from the hash grid. The non-linear MLP explicitly mixes inter-dimensional cues, and joint training with the entropy loss preserves only the bits-relevant directions of $\bm{f}^h$. This enables the hash grid to grow its representational capacity freely without inflating entropy-parameter cost, while the entropy context becomes structured and low-dimensional.

\subsection{Adaptive residual entropy coding}
\label{subsec:delta_feature_coding}

\subsubsection{Adaptive residual quantization}

Prior methods~\cite{HAC, chen2025hac++, Contextgs} entropy-code anchor attributes directly, ignoring the much stronger regularity of residuals around spatial-context-predicted means. We reduce entropy by encoding residuals of anchor attributes $\mathcal{A}$ with respect to their context-predicted means. For the $i$-th anchor $\bm{x}^a_i$, let $f_i \in \{ f^a_i, l_i, o_i \}$ denote any component of $\mathcal{A}_i$. To operate on the discrete symbol space for entropy coding, we quantize the anchor attribute residual $\Delta f_i$ with adaptive residual quantization. Specifically, we first predict the quantization step adjustment $\Delta q_i$ and the entropy model parameters $\bm{\mu}^\text{s}_i,\ \bm{\sigma}^\text{s}_i,\ \bm{\pi}^\text{s}_i$ for $f_i$ from the fused context $\bm{f}^c_i$.
\begin{equation}
\begin{aligned}
    \Delta q_i &= \mathrm{MLP_q}(\bm{f}^c_i), \\
    \bm{\mu}^\text{s}_i,\ \bm{\sigma}^\text{s}_i,\ \bm{\pi}^\text{s}_i &= \mathrm{MLP_c}(\bm{f}^c_i).
\end{aligned}
\end{equation}

Then, we dynamically quantize the anchor attribute residual $\Delta f_i$ in terms of the predicted mean $\bm{\mu}^s_i$:
\begin{equation}
\begin{aligned}
\Delta \bm{f}_i &= \bm{f}_i - \bm{\mu}^s_i, \qquad
\bm{r}_i = \left\lfloor \Delta \bm{f}_i \oslash \bm{q}_i \right\rceil \in \mathbb{Z}^{D^f}, \\
\Delta \hat{\bm{f}}_i &= \bm{r}_i \odot \bm{q}_i, \qquad
\hat{\bm{f}}_i = \bm{\mu}^s_i + \Delta \hat{\bm{f}}_i ,
\end{aligned}
\end{equation}
where $\odot$ and $\oslash$ denote element-wise product and division, $\bm{q}_i = q_0 \cdot \bigl(1 + \tanh(\Delta q_i)\bigr)$, and $q_0$ is the attribute-specific initial quantization step size. $\bm{r}_i$ is the integer symbol encoded by the arithmetic coder, $\Delta\hat{\bm{f}}_i$ the quantized residual, and $\hat{\bm{f}}_i$ the reconstruction. Decoding requires only $\bm{r}_i$ and $\bm{f}^c_i$, which predicts $\bm{\mu}^s_i$ and $\bm{q}_i$. During training, we quantize by injecting uniform noise with bin width $q_i$: $\Delta \hat{\bm{f}}_i^{\text{train}} = \Delta f_i + \boldsymbol{\epsilon}_i$, where $\boldsymbol{\epsilon}_i \sim
\mathcal{U}\!\left(-\tfrac{q_i}{2}, \tfrac{q_i}{2}\right)$. This yields the straight-through gradient $\frac{\partial \Delta \hat{\bm{f}}_i^{\text{train}}}{\partial \Delta f_i} = \mathbf{I}$, enabling gradient backpropagation. Compared with direct quantization, centering on $\bm{\mu}^\text{s}_i$ contracts the dynamic range and lowers residual entropy, which improves entropy coding efficiency.

\subsubsection{Residual distribution modeling}
For an anchor $\bm{x}^a_i$ and its quantized anchor attribute residual $\Delta \hat{\bm{f}}_i\in\mathbb{R}^{D^f}$, the probability of $\Delta \hat{\bm{f}}_i$ is computed using the cumulative distribution function difference over the quantization bins:
\begin{equation}
\begin{aligned}
    p(\Delta \hat{\bm{f}}_i)
    &= \prod_{j=1}^{D^f} \int_{\Delta \hat{f}_{i,j}-\frac{q_{i,j}}{2}}^{\Delta \hat{f}_{i,j}+\frac{q_{i,j}}{2}}
       \phi\!\left(x; 0, \sigma^\text{s}_{i,j}\right)\,dx \\
    &= \prod_{j=1}^{D^f} \Bigl[
       \Phi\!\left(\Delta \hat{f}_{i,j}+\tfrac{q_{i,j}}{2}; 0, \sigma^\text{s}_{i,j}\right)
       - \Phi\!\left(\Delta \hat{f}_{i,j}-\tfrac{q_{i,j}}{2}; 0, \sigma^\text{s}_{i,j}\right)
       \Bigr],
\end{aligned}
\end{equation}
where $\phi$ and $\Phi$ are the probability density function and the cumulative distribution function of a zero-mean Gaussian distribution, since the predicted residual $\Delta \hat{\bm{f}}_i$ is centered near zero. We then entropy code $p(\Delta \hat{\bm{f}}_i)$ using arithmetic coding \cite{witten1987arithmetic} and store the resulting bytestream payload of the compressed probability symbols. With per-element code length $-\log_2 p(\Delta \hat{f}_{i,j})$, this approach substantially reduces the storage footprint and the coding entropy.

\subsection{Causal entropy model refinement}

To further improve bitrate efficiency and reduce the parameter redundancies of the entropy model, we introduce a FiLM-modulated autoregressive GRU context model that refines the entropy model parameters by exploiting the causal relations inside the anchor feature groups and operating on the recurrent context hidden states. For each anchor $\bm{x}_i^a$, we split its anchor feature $\bm{f}^a_i$ into $N^C$ chunks $\bm{f}^a_i =[\bm{f}^a_{i,1};\dots;\bm{f}^a_{i,N^C}]$, where the chunk index $n\in\{1,\dots,N^C\}$ follows the decoding order and each chunk contains $C$ contiguous feature channels. Then, we autoregressively predict the distribution parameters $\mu^C_{i,n}, \sigma^C_{i,n}, \pi^C_{i,n}$ for $\bm{f}^a_{i,n}$, which are utilized to compute the distribution parameters for $\bm{f}^a_i$ as follows:
\begin{equation}
\mu^C_{i}, \sigma^C_{i}, \pi^C_{i} = \bigoplus_{n=1}^{N^C} \left( \mu^C_{i,n}, \sigma^C_{i,n}, \pi^C_{i,n} \right).
\end{equation}

Existing methods~\cite{zhan2025cat3dgs, chen2025hac++} build per-chunk causal context by concatenating raw decoded chunks, which causes the predictor input to scale linearly with the chunk count; higher-order interactions are captured only implicitly through the predictor's first MLP layer. In contrast, our method has constant per-step input cost, explicit gating-based modeling of inter-chunk interactions, and parameter-efficient chunk specialization.

\subsubsection{Recurrent autoregressive GRU context model}
We refine the chunkwise entropy parameters by conditioning on previously decoded feature groups within the same anchor. Let $\hat{\bm{f}}^a_{i,<n}=[\hat{\bm{f}}^a_{i,1};\dots;\hat{\bm{f}}^a_{i,n-1}]$ denote the already decoded chunks. We first form a base context vector $\bm{s}_i = \bigl[\bm{\mu}^s_i;\bm{\sigma}^s_i;\bm{\pi}^s_i\bigr]$ by concatenating the per-channel base statistics predicted from $\bm{f}^c_i$,
and encode it into an initial hidden state:
\begin{equation}
\bm{h}^0_i = \mathrm{MLP}_{\mathrm{cxt}}(\bm{s}_i).
\end{equation}

To inject causal information, we embed each previously decoded chunk into a token sequence.
Let $\bm{z}_{i,t}=\mathrm{MLP}_{\mathrm{prev}}(\hat{\bm{f}}^a_{i,t})$ for $t<n$, and update a single-layer GRU by exactly one step per chunk:
\begin{equation}
\bm{h}^{1}_i = \bm{h}^0_i, \qquad
\bm{h}^{n}_i = \mathrm{GRU}\!\left(\bm{z}_{i,n-1},\ \bm{h}^{n-1}_i\right), \quad n=2,\dots,N^C,
\label{eq:gru}
\end{equation}
where $\bm{h}^{n}_i$ is the chunk-specific latent summarizing all preceding decoded chunks. The cached hidden state advances through one constant-cost GRU step per chunk, without reprocessing decoded chunks. Operating on the recurrent hidden state $\bm{h}^{n}_i$ provides a compact summary of all previously decoded chunks, allowing the model to capture higher-order intra-anchor dependencies. Unlike in existing methods, the size of $\bm{h}^{n}_i$ is independent of $n$, so the predictor input cost is constant per chunk. Also, the GRU's gating models multiplicative inter-chunk interactions explicitly rather than relying on the predictor's first MLP layer to discover them.

\subsubsection{Anchor-conditional FiLM modulation}
While $\bm{h}^{n}_i$ aggregates causal tokens, its optimal scale and bias depend on the anchor-specific base statistics $\bm{s}_i$. Directly feeding $\bm{h}^{n}_i$ into the parameter prediction model can be restrictive, since the same hidden magnitude and offset must simultaneously fit anchors with very different predicted uncertainty patterns and mixture preferences. To provide an efficient anchor-adaptive reparameterization, we apply feature-wise linear modulation (FiLM)~\cite{perez2018film} to the chunk latent. We predict per-channel modulation vectors from $\bm{h}^{n}_i$,
\begin{equation}
\bm{\gamma}^{n}_i,\bm{\beta}^{n}_i
= \mathrm{MLP}_{\mathrm{FiLM}}\!\left(\bm{h}^{n}_i\right),
\end{equation}
and modulate a copy of the hidden representation as:
\begin{equation}
\tilde{\bm{h}}^{n}_i
= \bm{h}^{n}_i \odot \left(\mathbf{1}+\tanh(\bm{\gamma}^{n}_i)\right) + \bm{\beta}^{n}_i.
\end{equation}
This modulation acts as a lightweight chunkwise conditioning layer that both improves optimization stability and increases the effective capacity of the refinement module without deepening the GRU, which preserves the parameter footprint and improves coding efficiency.

\subsubsection{Weight-shared chunk-positional predictor}
To enhance sample efficiency and minimize model size, we utilize a shared parameter prediction model across all feature chunks. However, different chunks $\bm{f}^a_{i,n}$ correspond to different channel subsets and exhibit distinct statistics. To enable the shared parameter model to specialize for each chunk without duplicating weights, we introduce a learnable group positional embedding $\bm{g}_{n} = \mathrm{Emb}(n)$. Conditioned on the modulated causal hidden state $\tilde{\bm{h}}^{n}_i$, the base parameter embedding $\bm{b}_{i,n}
=
\mathrm{MLP}_{\mathrm{base}}\!\left(\bm{\mu}^s_{i,n}\ ,\ \bm{\sigma}^s_{i,n}\ ,\ \bm{\pi}^s_{i,n}\right)
$, and the group positional embedding $\bm{g}_{n}$, we apply the shared parameter model to predict the distribution parameters for $\bm{f}^a_{i,n}$:
\begin{equation}
\mu^C_{i,n}, \sigma^C_{i,n}, \pi^C_{i,n}=\mathrm{MLP}_{\mathrm{p}}(\tilde{\bm{h}}^{n}_i \oplus \bm{b}_{i,n} \oplus \bm{g}_{n}).
\end{equation}

Here, $\bm{\mu}^C_{i,n}$ and $\bm{\sigma}^C_{i,n}$ parameterize the refined Gaussian distribution component for channels in chunk $n$, while $\bm{\pi}^C_{i,n}$ is the mixture logit that controls the probability mass assigned to the refined component (Eq.~\eqref{eq:mixture}).

\subsubsection{Residual mixture distribution with refined context}
To improve the probability estimation accuracy of the anchor feature residual $\Delta\bm{f}^a_i$, we leverage a Gaussian mixture distribution that combines the base entropy model parameters predicted from the hash context feature $\bm{f}^c_i$ and the refinement entropy model parameters predicted from the causal context relations inside the anchor feature $\bm{f}^a_i$. The resulting mixture probability distribution is defined as:
\begin{equation}
\begin{aligned}
p(\Delta \hat{\bm{f}}^{a}_i)
= \prod_{j=1}^{D^f}\ \sum_{k\in\{\mathrm{s},\mathrm{c}\}} \pi^{k}_{i,j}
\Bigl[
\Phi\!\left(\Delta \hat{f}^{a}_{i,j}+\tfrac{q_{i,j}}{2};\,\mu^{k}_{i,j},\,\sigma^{k}_{i,j}\right)
-\Phi\!\left(\Delta \hat{f}^{a}_{i,j}-\tfrac{q_{i,j}}{2};\,\mu^{k}_{i,j},\,\sigma^{k}_{i,j}\right)
\Bigr].
\end{aligned}
\label{eq:mixture}
\end{equation}

This mixture entropy model enables the base predictor to capture coarse residual statistics from $\bm{f}^c_i$, while the refinement component adjusts the probability mass using causal dependencies within anchor feature groups, improving probability accuracy and bitrate efficiency.

\subsection{Optimization}

\subsubsection{Residual entropy loss}
To improve the estimation accuracy of $p(\Delta \hat{\bm{f}}_i)$, we minimize the predicted entropy of the anchor attribute residuals. Therefore, we define the residual entropy loss as the total expected bit consumption over all $N$ anchors:
\begin{equation}
\mathcal{L}_{\Delta \mathrm{entropy}}
=
\sum_{i=1}^{N} m_i^{a}
\Bigl[
\sum_{\bm{f}\in\{\bm{f}^{a},\bm{l}\}}
\sum_{j=1}^{D^{f}}
-\log_{2} p\!\left(\Delta\hat{f}_{i,j}\right)
+
\sum_{k=1}^{K} m_{i,k}
\sum_{j=1}^{3}
-\log_{2} p\!\left(\Delta\hat{o}_{i,k,j}\right)
\Bigr],
\end{equation}
where $K$ is the number of offset vectors attached to each anchor, and $m_i^{a}$ and $m_{i,k}$ are learnable binary masks \cite{Lee} used to prune inactive anchors indexed by $i$ and their $k$-th offsets, respectively.

\subsubsection{Training process}
During training, we jointly optimize the rendering losses and the residual entropy loss, improving rendering quality while constraining the overall bitrate. The total training loss is defined as:
\begin{equation}
\mathcal{L}_{\mathrm{total}}
=
(1-\lambda_{\mathrm{SSIM}})\,\mathcal{L}_{1}
+
\lambda_{\mathrm{SSIM}}\,\mathcal{L}_{\mathrm{SSIM}}
+
\lambda_{\mathrm{vol}}\,\mathcal{L}_{\mathrm{vol}}
+
\frac{\lambda_{\mathrm{r}}}{N N_p}
(\mathcal{L}_{\Delta \mathrm{entropy}} + \mathcal{L}_{\mathrm{hash}}),
\end{equation}
where $\lambda_{\mathrm{r}}$ is the rate weight for balancing the bitrate trade-off, $N_p$ is the number of coded parameters per anchor, $\mathcal{L}_{\mathrm{hash}}$ is the bit consumption of the binarized hash grid \cite{chen2025hac++}, and $\mathcal{L}_{\mathrm{vol}}
=
\frac{1}{N_{\mathrm{vis}}}
\sum_{i=1}^{N_{\mathrm{vis}}}
\prod_{c=1}^{3} s_{i,c}$ \cite{scaffold} is the mean volume per Gaussian for $N_{\mathrm{vis}}$ visible Gaussians.

%% file: sec/4_experiments.tex
\section{Experiments}
\label{sec:experiments}

\input{tables/quant}

\begin{figure}[!t]
  \centering
  \includegraphics[width=1\columnwidth]{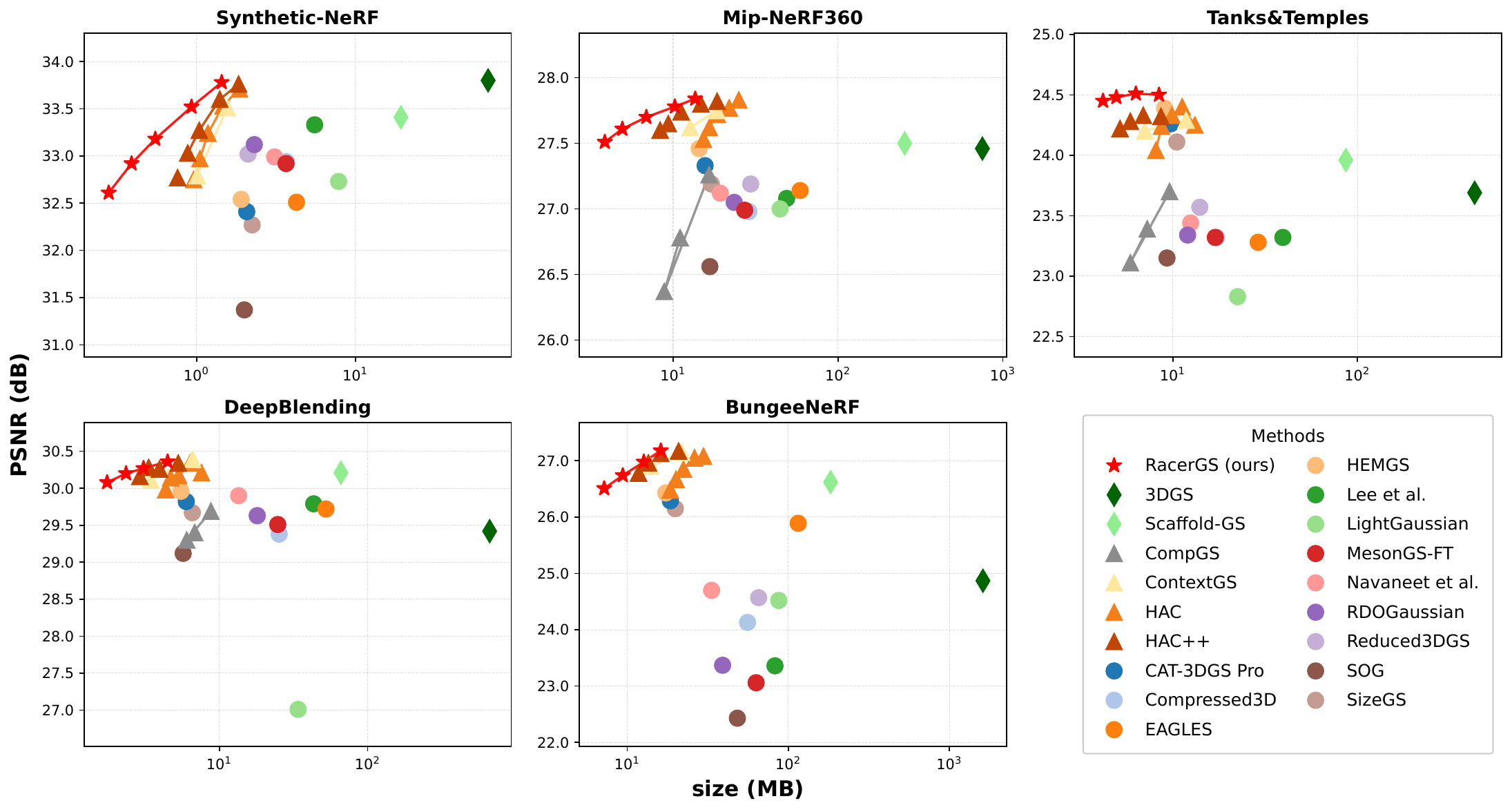}
  \caption{\textbf{RD curve comparisons.} RacerGS achieves better RD performance than the baseline methods across datasets.}
  \label{fig_quant}
\end{figure}

\begin{figure}[!t]
  \centering
  \includegraphics[width=1\columnwidth]{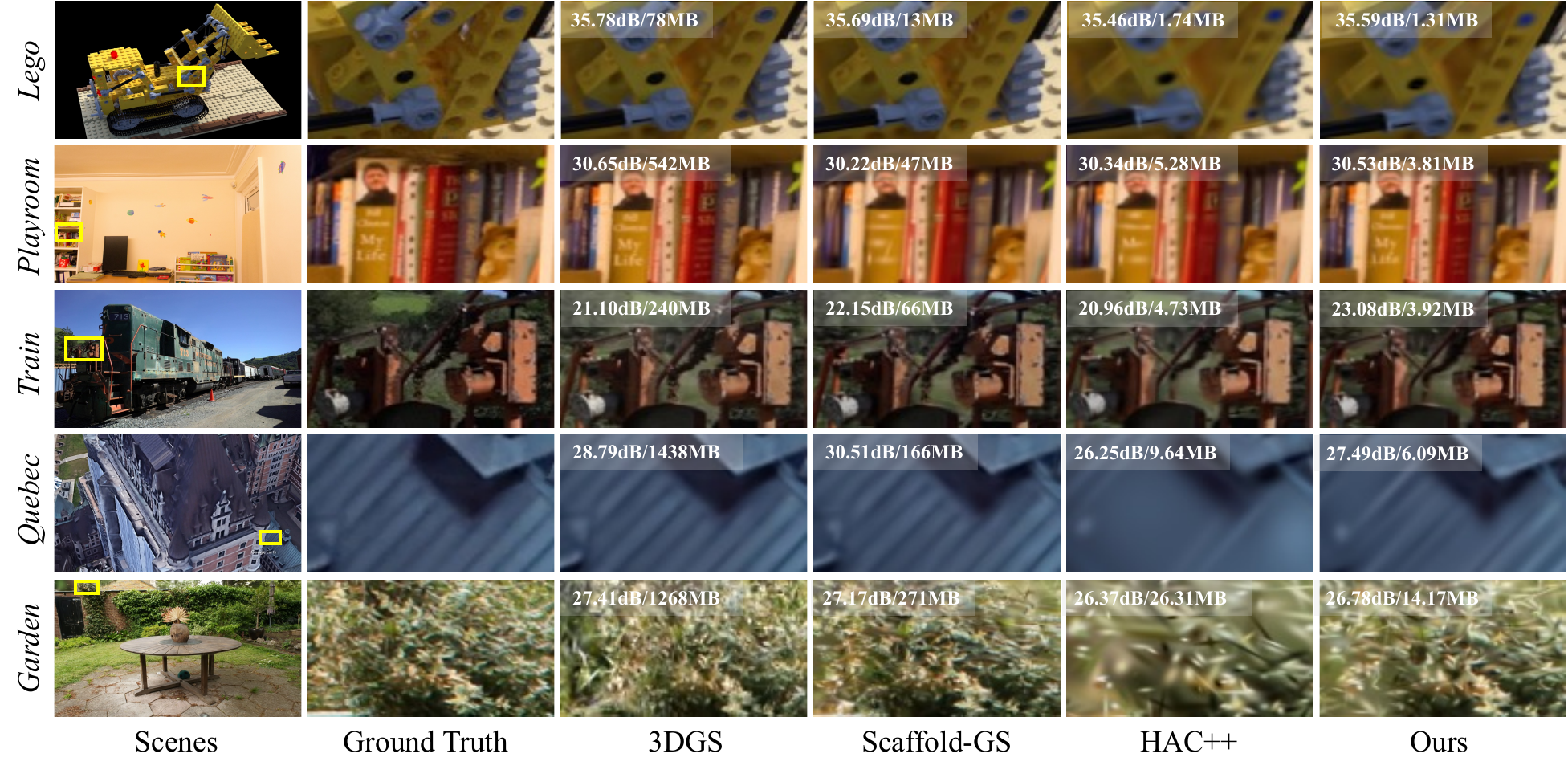}
  \caption{\textbf{Qualitative comparisons.} RacerGS preserves details better and produces fewer artifacts than the baseline methods across scenes from diverse datasets.}
  \label{fig_qua}
\end{figure}

\subsection{Implementation}
\label{subsec:implementation}

\noindent\textbf{Datasets.}
We systematically conduct experiments on five comprehensive datasets, including one synthetic dataset, Synthetic-NeRF~\cite{NeRF}, and four real-world datasets: BungeeNeRF~\cite{BungeeNeRF}, DeepBlending~\cite{deepblending}, Mip-NeRF360~\cite{mip360}, and Tanks\&Temples~\cite{tant}.

\noindent\textbf{Metrics.}
We evaluate rendering quality with PSNR, SSIM~\cite{ssim}, and LPIPS~\cite{LPIPS}, and use storage size (MB) as the bitrate measure. Image metrics are averaged over test views and then over scenes in each dataset. Reported sizes include coded anchor positions, attributes, the hash grid, and masks, plus decoder-side network parameters and bounds; Table~\ref{tab:total_param_size} gives the component breakdown. RD curves are obtained by varying the trade-off coefficient $\lambda_r$. We also report training, encoding, and decoding times, peak GPU memory, and rendering speed in FPS.

\noindent\textbf{Baselines.}
We compare RacerGS with a diverse set of 3D Gaussian compression baselines, including codebook and pruning methods~\cite{Lee, Simon, Navaneet, Lightgaussian, Reduced3DGS, RDOGaussian, MesonGS, liu2024hemgs, zhan2025cat3dgspro, xie2025sizegs}, entropy- and sorting-based approaches~\cite{EAGLES, SOG}, and anchor-based methods~\cite{scaffold, chen2025hac++, HAC, liu2024compgs, Contextgs}.

\noindent\textbf{Implementation details.}
All experiments are run on a single NVIDIA RTX A6000 GPU. Following prior work, we train for $30{,}000$ iterations. We obtain different bitrate operating points by varying the rate weight $\lambda_r$ from $4\times10^{-3}$ (low rate) to $0.5\times10^{-3}$ (high rate; $1\times10^{-3}$ on Tanks\&Temples, DeepBlending, and BungeeNeRF). We set $\lambda_{\mathrm{SSIM}}=0.2$ and $\lambda_{\mathrm{vol}}=0.001$. The anchor feature dimension is $D^a=50$, with $K=10$ offsets per anchor. For causal entropy refinement, each anchor feature is split into $N^C=5$ chunks. The base quantization step $q_0$ is set to $1$, $0.001$, and $0.2$ for anchor features $\bm{f}^a$, scaling $\bm{l}$, and offsets $\bm{o}$, respectively.

\begin{table}[!t]
    \small
    \centering
\caption{\textbf{Storage sizes (MB) of different components} on Mip-NeRF360 (see Appendix~\ref{sec:appendix} for results with additional $\lambda_r$ values).}
    \begin{tabular}{ccccccccc}
        \toprule[1.5pt]
        $\lambda_r$   & \textbf{Total Size}   & $\bm{x}^a$ & $\bm{f}^a$ & $\bm{l}$ & $\bm{o}$ & $\mathcal{H}$ & $\bm{m}$ & MLP \\ \midrule
        $4e\text{-}3$   & 3.85  &0.19 &1.89 &0.80 &0.48 &0.05 &0.10 &0.34 \\
        $0.5e\text{-}3$ & 13.61 &0.44 &8.15 &2.44 &1.83 &0.14 &0.27 &0.34 \\
        \bottomrule[1.5pt]
    \end{tabular}
    \label{tab:total_param_size}
\end{table}

\begin{table}[!t]
\centering
\small

\begin{minipage}[t]{0.49\columnwidth}
\centering
\caption{\textbf{Encoding and decoding times} across all datasets.}
\footnotesize
\setlength{\tabcolsep}{3pt}
\begin{tabular}{ccccc}
    \toprule
    \multirow{2}{*}{\textbf{Datasets}}
    & \multicolumn{2}{c}{\textbf{Enc. Time (s)}}
    & \multicolumn{2}{c}{\textbf{Dec. Time (s)}} \\ \cmidrule(lr){2-3}\cmidrule(lr){4-5}
    & low rate & high rate & low rate & high rate \\
    \midrule
    Synthetic-NeRF   & 0.63 & 1.28 & 0.89 & 1.75 \\
    Mip-NeRF360      & 4.24 & 8.59 & 7.40 & 14.20 \\
    Tanks\&Temples   & 4.53 & 7.48 & 7.80 & 12.63 \\
    DeepBlending     & 1.85 & 3.01 & 4.03 & 6.25 \\
    BungeeNeRF       & 10.03 & 16.26 & 16.43 & 25.98 \\
    \bottomrule
\end{tabular}

\label{tab:enc_dec_time_all}
\end{minipage}
\hfill
\begin{minipage}[t]{0.49\columnwidth}
\centering
\caption{\textbf{Training time, peak GPU memory, and FPS} on Mip-NeRF360.}
\footnotesize
\setlength{\tabcolsep}{4pt}
\renewcommand{\arraystretch}{1.35}
\begin{tabular}{lccc}
    \toprule
    \textbf{Method} & \textbf{Train (s)}$\downarrow$ & \textbf{Mem. (GB)}$\downarrow$ & \textbf{FPS}$\uparrow$ \\
    \midrule
    3DGS        & 1607 & -- & 102 \\
    Scaffold-GS & \textbf{1350} & 15.8 & 142 \\
    HAC++       & 2447 & 17.4 & 146 \\
    \midrule
    \textbf{RacerGS} & 2850 & \textbf{13.9} & \textbf{171} \\
    \bottomrule
\end{tabular}

\label{tab:train_time_fps_compare}
\end{minipage}
\end{table}

\subsection{Experimental results}
\label{subsec:results}

\noindent\textbf{Quantitative.}
As shown in Table~\ref{tab:main_quantitative}, RacerGS achieves highly competitive compression while preserving rendering quality across all datasets. On average, it provides over \textbf{$\mathbf{150}\boldsymbol{\times}$} compression over vanilla 3DGS~\cite{3DGS} and \textbf{$\mathbf{28}\boldsymbol{\times}$} over Scaffold-GS~\cite{scaffold}, while outperforming recent codebook/pruning~\cite{Lee, Simon, Navaneet, Lightgaussian, Reduced3DGS, RDOGaussian, MesonGS}, entropy/sorting-based~\cite{EAGLES, SOG}, and anchor-based~\cite{chen2025hac++, HAC, liu2024compgs} methods at smaller model sizes. At the high-rate Mip-NeRF360 point, it uses 13.61\,MB versus HAC++'s 18.48\,MB, with a PSNR of 27.84 versus 27.82\,dB. Fig.~\ref{fig_quant} shows the RD behavior across the rate sweep. These gains come from the hash projection model, residual quantization, and FiLM-modulated autoregressive GRU context modeling, which jointly improve context representation and entropy estimation.

\noindent\textbf{Qualitative.}
As shown in Fig.~\ref{fig_qua}, RacerGS produces higher visual fidelity across diverse scenes, with fewer leaf artifacts in \textit{garden}, clearer book title details in \textit{playroom}, and sharper metal structures in \textit{train}. Even at comparable or smaller sizes, it better preserves edges and material details while keeping backgrounds smooth. These comparisons highlight its ability to retain fine structures and textured regions under tight storage constraints.

\noindent\textbf{Efficiency analysis.}
As shown in Table~\ref{tab:train_time_fps_compare}, RacerGS achieves the fastest rendering speed of \textbf{171 FPS} on Mip-NeRF360 and the lowest peak GPU memory. Encoding and decoding remain efficient across datasets, taking \textbf{0.63\,s}--\textbf{16.26\,s} and \textbf{0.89\,s}--\textbf{25.98\,s}, respectively (Table~\ref{tab:enc_dec_time_all}). Table~\ref{tab:total_param_size} further shows that storage is dominated by anchor features, while the hash grid and MLP modules add only minor overhead, indicating effective entropy control and smooth bitrate allocation. Although RacerGS takes longer to train than the compared baselines, it offers lower peak memory usage and faster rendering, providing a favorable trade-off for applications with repeated scene rendering.

\subsection{Ablation study}
\label{subsec:ablation}
\input{tables/ablation}

\input{tables/context}

\noindent\textbf{Hash projection model.}
Removing the hash projection model degrades both quality and compression: PSNR drops from 27.84 to 27.62, and model size grows from 13.61 MB to 15.21 MB. Projecting multi-dimensional hash encodings into compact fused context features provides a stronger spatial prior and complementary cues for entropy parameter prediction. A single linear projection captures only part of the gain (27.76\,dB / 14.28\,MB). The context-dimension sweep in Appendix Table~\ref{tab:extra_analyses} varies $D^c$ from 12 to 48 and changes size by at most 0.35\,MB and PSNR by at most 0.04\,dB, suggesting limited sensitivity within that range. Together, these results suggest that learning a structured context representation contributes more to the gains than the precise choice of context dimensionality.

\noindent\textbf{Residual quantization.}
Without residual quantization, model size rises sharply from 13.61 MB to 18.24 MB, and PSNR slightly decreases, showing that centering anchor attributes around the predicted means effectively reduces residual entropy and dynamic range for more efficient entropy coding. The anchor-feature symbol analysis in Appendix Table~\ref{tab:extra_analyses} is consistent with that result: prediction-centered symbols have an empirical entropy of 5.22 rather than 8.27 bits per element, and the arithmetic-coded cost falls from 8.44 to 5.38 bits per element. The reduction in both quantities confirms that the lower symbol entropy translates into practical coding savings while preserving reconstruction quality.

\noindent\textbf{Autoregressive GRU context model.}
Replacing the GRU with HAC++-style decoded-chunk concatenation under the same pipeline increases size from 13.61 to 16.87\,MB at nearly equal PSNR (Table~\ref{tab:context_controlled}, Table~\ref{tab:ablation}). In Table~\ref{tab:context_chunks}, both cost 68 bits per anchor for the first chunk, before feature history is available. By the fifth chunk, the GRU costs 39.0 versus 80.1 bits per anchor for concatenation, consistent with a greater benefit from accumulated context. Removing FiLM raises size to 17.32\,MB (Table~\ref{tab:ablation}), worse even than concatenation; recurrence and modulation work together in this configuration. By summarizing previously decoded feature groups through recurrent hidden states, the GRU supplies additional conditional information that lowers coding entropy and improves RD performance.

%% file: tables/quant.tex
\begin{table*}[t!]
  \centering
  \small
\caption{\textbf{Quantitative comparisons.} Storage sizes are measured in MB. The best and second-best compression results are highlighted in \colorbox{green!25}{green} and \colorbox{blue!8}{blue}, respectively. Overall, RacerGS outperforms the baselines in size reduction and fidelity preservation across diverse datasets.}
  \setlength{\tabcolsep}{5.7pt}
\begingroup\scriptsize
  \setlength{\aboverulesep}{0pt}\setlength{\belowrulesep}{0pt}
  \begin{minipage}{\textwidth}
    \centering
    \begin{tabular}{llcccccccccccc}
      \toprule[1pt]
      \multicolumn{2}{l}{\textbf{Datasets}} &
        \multicolumn{4}{c}{\textbf{Synthetic-NeRF}} &
        \multicolumn{4}{c}{\textbf{Mip-NeRF360}} &
        \multicolumn{4}{c}{\textbf{Tanks\&Temples}} \\
      \cmidrule(lr){3-6}\cmidrule(lr){7-10}\cmidrule(lr){11-14}
    \rowcolor{gray!10}
      \multicolumn{2}{l}{\textbf{Methods}} &
        psnr$\uparrow$ & ssim$\uparrow$ & lpips$\downarrow$ & size$\downarrow$ &
        psnr$\uparrow$ & ssim$\uparrow$ & lpips$\downarrow$ & size$\downarrow$ &
        psnr$\uparrow$ & ssim$\uparrow$ & lpips$\downarrow$ & size$\downarrow$ \\
      \midrule
      \multicolumn{2}{l}{\textbf{3DGS}}&33.80&0.970&0.031&68.46&27.46&0.812&0.222&750.9&23.69&0.844&0.178&431.0\\
      \multicolumn{2}{l}{\textbf{Scaffold-GS}}&33.41&0.966&0.035&19.36&27.50&0.806&0.252&253.9&23.96&0.853&0.177&86.50\\ \midrule
      \multicolumn{2}{l}{\textbf{Lee~\etal}}&33.33&0.968&\cellcolor{blue!8}{0.034}&5.54&27.08&0.798&0.247&48.80&23.32&0.831&0.201&39.43\\
      \multicolumn{2}{l}{\textbf{Compressed3D}}&32.94&0.967&\cellcolor{green!25}{0.033}&3.68&26.98&0.801&0.238&28.80&23.32&0.832&0.194&17.28\\
      \multicolumn{2}{l}{\textbf{EAGLES}}&32.51&0.964&0.039&4.26&27.14&\cellcolor{blue!8}{0.809}&\cellcolor{blue!8}{0.231}&58.91&23.28&0.835&0.203&28.99\\
      \multicolumn{2}{l}{\textbf{SizeGS}}&32.27&0.956&0.043&2.24&27.19&0.788&0.253&17.05&24.11&0.833&0.191&10.51\\
      \multicolumn{2}{l}{\textbf{LightGaussian}}&32.73&0.965&0.037&7.84&27.00&0.799&0.249&44.54&22.83&0.822&0.242&22.43\\
      \multicolumn{2}{l}{\textbf{SOG}}&31.37&0.959&0.043&2.00&26.56&0.791&0.241&16.70&23.15&0.828&0.198&9.30\\
      \multicolumn{2}{l}{\textbf{Navaneet~\etal}}&32.99&0.966&0.037&3.10&27.12&0.806&0.240&19.33&23.44&0.838&0.198&12.50\\
      \multicolumn{2}{l}{\textbf{Reduced3DGS}}&33.02&0.967&0.035&2.11&27.19&0.807&\cellcolor{green!25}{0.230}&29.54&23.57&0.840&0.188&14.00\\
      \multicolumn{2}{l}{\textbf{CAT-3DGS Pro}}&32.41&0.959&0.041&2.07&27.33&0.793&0.245&15.62&24.26&0.838&0.188&9.74\\
      \multicolumn{2}{l}{\textbf{RDOGaussian}}&33.12&0.967&0.035&2.31&27.05&0.802&0.239&23.46&23.34&0.835&0.195&12.03\\
      \multicolumn{2}{l}{\textbf{HEMGS}}&32.54&0.962&0.038&1.91&27.46&0.797&0.241&14.38&24.39&0.842&0.185&9.03\\
      \multicolumn{2}{l}{\textbf{MesonGS-FT}}&32.92&0.968&\cellcolor{green!25}{0.033}&3.66&26.99&0.796&0.247&27.16&23.32&0.837&0.193&16.99\\
      \midrule
      \multicolumn{2}{l}{\textbf{HAC (lr)}}&33.24&0.967&0.037&1.18&27.53&0.807&0.238&15.26&24.04&0.846&0.187&8.10\\
      \multicolumn{2}{l}{\textbf{HAC (hr)}}&33.71&0.968&\cellcolor{blue!8}{0.034}&1.86&27.77&\cellcolor{green!25}{0.811}&\cellcolor{green!25}{0.230}&21.87&24.40&0.853&\cellcolor{blue!8}{0.177}&11.24\\
      \multicolumn{2}{l}{\textbf{ContextGS (lr)}}&32.79&0.965&0.040&1.01&27.62&0.808&0.237&12.68&24.20&0.852&0.184&7.05\\
      \multicolumn{2}{l}{\textbf{ContextGS (hr)}}&33.51&0.968&0.035&1.56&27.75&\cellcolor{green!25}{0.811}&\cellcolor{blue!8}{0.231}&18.41&24.29&\cellcolor{green!25}{0.855}&\cellcolor{green!25}{0.176}&11.80\\
      \multicolumn{2}{l}{\textbf{CompGS (lr)}}&-&-&-&-&26.37&0.778&0.267&8.83&23.11&0.815&0.236&5.89\\
      \multicolumn{2}{l}{\textbf{CompGS (hr)}}&-&-&-&-&27.26&0.803&0.239&16.50&23.70&0.837&0.208&9.60\\
      \multicolumn{2}{l}{\textbf{HAC++ (lr)}}&33.03&0.966&0.039&\cellcolor{blue!8}{0.88}&27.60&0.803&0.253&\cellcolor{blue!8}{8.34}&24.22&0.849&0.190&\cellcolor{blue!8}{5.18}\\
      \multicolumn{2}{l}{\textbf{HAC++ (hr)}}&\cellcolor{blue!8}{33.76}&\cellcolor{blue!8}{0.969}&\cellcolor{green!25}{0.033}&1.84&\cellcolor{blue!8}{27.82}&\cellcolor{green!25}{0.811}&\cellcolor{blue!8}{0.231}&18.48&24.32&\cellcolor{blue!8}{0.854}&0.178&8.63\\
      \midrule
      \multicolumn{2}{l}{\textbf{RacerGS (lr)}}&32.61&0.964&0.041&\cellcolor{green!25}{0.28}&27.51&0.801&0.262&\cellcolor{green!25}{3.85}&\cellcolor{blue!8}{24.45}&0.847&0.195&\cellcolor{green!25}{4.19}\\
      \multicolumn{2}{l}{\textbf{RacerGS (hr)}}&\cellcolor{green!25}{33.78}&\cellcolor{green!25}{0.970}&0.035&1.44&\cellcolor{green!25}{27.84}&\cellcolor{green!25}{0.811}&0.243&13.61&\cellcolor{green!25}{24.50}&\cellcolor{green!25}{0.855}&0.183&8.42\\
      \bottomrule[1pt]
    \end{tabular}
  \end{minipage}

  \vspace{8pt}

  \begin{minipage}{\textwidth}
    \centering
    \begin{tabular}{llcccccccc}
      \toprule[1pt]
      \multicolumn{2}{l}{\textbf{Datasets}} &
        \multicolumn{4}{c}{\textbf{DeepBlending}} &
        \multicolumn{4}{c}{\textbf{BungeeNeRF}} \\
      \cmidrule(lr){3-6}\cmidrule(lr){7-10}
        \rowcolor{gray!10}
      \multicolumn{2}{l}{\textbf{Methods}} &
        psnr$\uparrow$ & ssim$\uparrow$ & lpips$\downarrow$ & size$\downarrow$ &
        psnr$\uparrow$ & ssim$\uparrow$ & lpips$\downarrow$ & size$\downarrow$ \\
      \midrule
      \multicolumn{2}{l}{\textbf{3DGS}}&29.42&0.899&0.247&663.9&24.87&0.841&0.205&1616\\
      \multicolumn{2}{l}{\textbf{Scaffold-GS}}&30.21&0.906&0.254&66.00&26.62&0.865&0.241&183.0\\ \midrule
      \multicolumn{2}{l}{\textbf{Lee~\etal}}&29.79&0.901&0.258&43.21&23.36&0.788&0.251&82.60\\
      \multicolumn{2}{l}{\textbf{Compressed3D}}&29.38&0.898&0.253&25.30&24.13&0.802&0.245&55.79\\
      \multicolumn{2}{l}{\textbf{EAGLES}}&29.72&0.906&\cellcolor{green!25}{0.249}&52.34&25.89&0.865&\cellcolor{blue!8}{0.197}&115.2\\
      \multicolumn{2}{l}{\textbf{SizeGS}}&29.67&0.898&0.257&6.57&26.15&0.831&0.251&19.88\\
      \multicolumn{2}{l}{\textbf{LightGaussian}}&27.01&0.872&0.308&33.94&24.52&0.825&0.255&87.28\\
      \multicolumn{2}{l}{\textbf{SOG}}&29.12&0.892&0.270&5.70&22.43&0.708&0.339&48.25\\
      \multicolumn{2}{l}{\textbf{Navaneet~\etal}}&29.90&0.907&\cellcolor{blue!8}{0.251}&13.50&24.70&0.815&0.266&33.39\\
      \multicolumn{2}{l}{\textbf{Reduced3DGS}}&29.63&0.902&\cellcolor{green!25}{0.249}&18.00&24.57&0.812&0.228&65.39\\
      \multicolumn{2}{l}{\textbf{CAT-3DGS Pro}}&29.82&0.901&0.255&5.98&26.29&0.836&0.248&18.52\\
      \multicolumn{2}{l}{\textbf{RDOGaussian}}&29.63&0.902&0.252&18.00&23.37&0.762&0.286&39.06\\
      \multicolumn{2}{l}{\textbf{HEMGS}}&29.96&0.904&0.253&5.51&26.43&0.840&0.241&17.34\\
      \multicolumn{2}{l}{\textbf{MesonGS-FT}}&29.51&0.901&\cellcolor{blue!8}{0.251}&24.76&23.06&0.771&0.235&63.11\\
      \midrule
      \multicolumn{2}{l}{\textbf{HAC (lr)}}&29.98&0.902&0.269&4.35&26.48&0.845&0.250&18.49\\
      \multicolumn{2}{l}{\textbf{HAC (hr)}}&30.34&0.906&0.258&6.35&27.08&0.872&0.209&29.72\\
      \multicolumn{2}{l}{\textbf{ContextGS (lr)}}&30.11&0.907&0.265&3.45&26.90&0.866&0.222&14.00\\
      \multicolumn{2}{l}{\textbf{ContextGS (hr)}}&\cellcolor{green!25}{30.39}&0.909&0.258&6.60&27.15&\cellcolor{blue!8}{0.875}&0.205&21.80\\
      \multicolumn{2}{l}{\textbf{CompGS (lr)}}&29.30&0.895&0.293&6.03&-&-&-&-\\
      \multicolumn{2}{l}{\textbf{CompGS (hr)}}&29.69&0.901&0.269&8.77&-&-&-&-\\
      \multicolumn{2}{l}{\textbf{HAC++ (lr)}}&30.16&0.907&0.256&\cellcolor{blue!8}{2.91}&26.78&0.858&0.235&\cellcolor{blue!8}{11.75}\\
      \multicolumn{2}{l}{\textbf{HAC++ (hr)}}&30.34&\cellcolor{blue!8}{0.911}&0.254&5.28&\cellcolor{blue!8}{27.17}&\cellcolor{green!25}{0.879}&\cellcolor{green!25}{0.196}&20.82\\
      \midrule
      \multicolumn{2}{l}{\textbf{RacerGS (lr)}}&30.08&0.906&0.271&\cellcolor{green!25}{1.75}&26.51&0.845&0.255&\cellcolor{green!25}{7.19}\\
      \multicolumn{2}{l}{\textbf{RacerGS (hr)}}&\cellcolor{blue!8}{30.36}&\cellcolor{green!25}{0.913}&0.256&4.48&\cellcolor{green!25}{27.18}&0.867&0.216&16.15\\
      \bottomrule[1pt]
    \end{tabular}
  \end{minipage}

  \endgroup
  \label{tab:main_quantitative}
\end{table*}

%% file: tables/ablation.tex
\begin{table}[!t]
    \centering
    \small
\caption{\textbf{Ablation studies} on the Mip-NeRF360. The best results are in \textbf{bold}.}
    \resizebox{\columnwidth}{!}{%
    \begin{tabular}{lcccccc}
        \toprule
        \textbf{Ablation Items} & \textbf{PSNR} $\uparrow$ & \textbf{SSIM} $\uparrow$ & \textbf{LPIPS} $\downarrow$ & \textbf{Size} (MB) $\downarrow$ & \textbf{FPS} $\uparrow$ & \textbf{Train (s)} $\downarrow$ \\
        \midrule
        W/o residual quantization & 27.83 & 0.810 & \textbf{0.242} & 18.24 & 146 & 2856 \\
        W/o hash projection  & 27.62 & 0.804 & 0.248 & 15.21 & 107 & 2865 \\
        Concat.\ context instead of GRU & 27.81 & 0.806 & 0.244 & 16.87 & 156 & \textbf{2841} \\
        GRU context w/o FiLM & 27.75 & 0.802 & 0.247 & 17.32 & 149 & 2854 \\
        \midrule
        \textbf{RacerGS (full)} & \textbf{27.84} & \textbf{0.811} & 0.243 & \textbf{13.61} & \textbf{171} & 2850 \\
        \bottomrule
    \end{tabular}
    }
    \label{tab:ablation}
\end{table}

%% file: tables/context.tex
\begin{table}[!t]
\centering
\small
\begin{minipage}[t]{0.4925\columnwidth}
\centering
\caption{Controlled comparison of intra-anchor context models on Mip-NeRF360.}
\label{tab:context_controlled}
\scriptsize
\setlength{\tabcolsep}{5.17pt}
\begin{tabular}{lccc}
\toprule
\textbf{Context model} & \textbf{Ctx params} & \textbf{PSNR}$\uparrow$ & \textbf{Size}$\downarrow$ \\
\midrule
No intra-anchor context & 0 & 27.69 & 18.20 \\
Concat.\ (HAC++-style) & $\approx$40k & 27.81 & 16.87 \\
Concat.\ (capacity-matched) & $\approx$48k & 27.82 & 15.24 \\
GRU w/o FiLM & $\approx$48k & 27.75 & 17.32 \\
\midrule
\textbf{GRU + FiLM (ours)} & $\approx$48k & \textbf{27.84} & \textbf{13.61} \\
\bottomrule
\end{tabular}
\end{minipage}
\hfill
\begin{minipage}[t]{0.4925\columnwidth}
\centering
\caption{Anchor-feature bits per anchor by chunk index $n$ on Mip-NeRF360.}
\label{tab:context_chunks}
\scriptsize
\setlength{\tabcolsep}{2.47pt}
\renewcommand{\arraystretch}{1.66}
\begin{tabular}{lcccccc}
\toprule
\textbf{Context model} & $n{=}1$ & $n{=}2$ & $n{=}3$ & $n{=}4$ & $n{=}5$ & \textbf{Total} \\
\midrule
No intra-anchor context & 68.0 & 80.5 & 86.0 & 90.5 & 95.5 & 420.5 \\
Concat.\ (HAC++-style) & 68.0 & 74.5 & 76.0 & 78.0 & 80.1 & 376.6 \\
\textbf{GRU + FiLM (ours)} & 68.0 & \textbf{61.0} & \textbf{54.0} & \textbf{47.0} & \textbf{39.0} & \textbf{269.0} \\
\bottomrule
\end{tabular}
\end{minipage}
\end{table}

%% file: sec/5_conclusion.tex
\section{Conclusion}
We presented RacerGS, a 3DGS compression framework that reduces storage while preserving rendering fidelity. RacerGS proposes compact hash-projected context features, adaptive residual quantization for lower entropy, and a FiLM-modulated autoregressive GRU context model to refine entropy estimates through causal anchor dependencies. Experiments and ablations demonstrate that RacerGS achieves substantially lower storage requirements than existing methods with comparable rendering quality across diverse datasets. Future work includes parallelizable context models and content-adaptive control to enable more efficient 3D rendering. By reducing the cost of storing and distributing 3D scenes, this work advances the goal of efficient visual computing: making richer representations of the world accessible with fewer resources.

%% file: sec/X_suppl.tex
\label{sec:appendix}

\section{Entropy coding}
Entropy coding is a lossless compression technique that represents discrete symbols with code lengths proportional to their information content. Given a discrete random variable $Y$ with probability mass function $p(y)$, the optimal code length $\ell$ for a symbol $y$ satisfies $\ell(y)\approx -\log_2 p(y)$, and the expected code length is bounded below by the Shannon entropy $\mathbb{E}[\ell(Y)] \ge H(Y) = -\sum_{y} p(y)\log_2 p(y)$ \cite{cover1999elements}. Entropy coders such as arithmetic coding \cite{witten1987arithmetic} achieve expected lengths arbitrarily close to $H(Y)$ when the model probability $p$ matches the true symbol distribution.

\section{Hash-grid-assisted design}
\label{sec:supp_hash}

HAC~\cite{HAC} introduces an anchor-based 3DGS method that reduces storage size by leveraging spatial consistency through hash features $\bm{f}^h=\mathrm{Interp}(\bm{x}^a, \mathcal{H})$ queried at the anchor position $\bm{x}^a$. The hash grid for dimension $d\in\{\mathrm{3D}, \mathrm{2D}\}$ is defined as $\mathcal{H}^{d}
=
\left\{\bm{\Theta}^{l}_{d}\in\mathbb{R}^{T^{l}_{d}\times d_h}\right\}_{l=1}^{L_{d}}$, where $\Theta$ is the hash embedding, $T$ is the table size, and $l$ indexes the $L$ multi-resolution levels. Although this hash-grid-assisted design achieves strong performance, substantial redundancy remains due to the spatial consistency of anchor attribute residuals and causal dependencies among anchor feature groups. We further exploit this spatial consistency and capture these causal dependencies through recurrent hidden states to produce a more compact 3DGS representation.

\section{Per-component coding time}

We analyze the per-component coding time of RacerGS on the Mip-NeRF360 dataset in Tables~\ref{tab:encoding_time} and~\ref{tab:decoding_time}. Overall, the total coding time increases as the rate weight $\lambda_r$ decreases, since a smaller $\lambda_r$ produces more anchors and representations with higher bitrates. The anchor feature component $\bm{f}^{a}$ dominates both encoding and decoding time, accounting for about $46\%\!\sim\!51\%$ of encoding time and $50\%\!\sim\!53\%$ of decoding time, because it contains the largest number of parameters and requires entropy modeling with the autoregressive GRU context model. The scaling parameters $\bm{l}$ and offsets $\bm{o}$ account for the next largest portion of the coding time, while the anchor positions $\bm{x}^a$ contribute only a small fraction of the overall computation. The remaining components introduce negligible overhead due to their compact representations. These results indicate that the computational cost is mainly concentrated in the entropy coding of anchor attributes, while the other components introduce minimal overhead, demonstrating that RacerGS maintains efficient coding complexity.

\section{Quantitative per-scene results}
\label{sec:supp_perscene}
Per-scene results for the Synthetic-NeRF dataset are shown in Table~\ref{tab:per_scene_syncnerf}. Per-scene results for the Mip-NeRF360 dataset are shown in Table~\ref{tab:per_scene_mipnerf}.
Per-scene results for the Tanks\&Temples dataset are shown in Table~\ref{tab:per_scene_tandt}.
Per-scene results for the DeepBlending dataset are shown in Table~\ref{tab:per_scene_deepblending}.
Per-scene results for the BungeeNeRF dataset are shown in Table~\ref{tab:per_scene_bungeenerf}.

\section{Additional analyses}
\input{tables/extra}

\begin{table}[!t]
\centering
\small
\caption{\textbf{Decoding time breakdown}. Arithmetic coding accounts for about 60\% of decoding time across all datasets, with the rest spent on network evaluation.}
\label{tab:dec_scaling}
\setlength{\tabcolsep}{7pt}
\renewcommand{\arraystretch}{1.08}
\begin{tabular}{lccc}
    \toprule
    \textbf{Dataset} & \textbf{Total (s)}
    & \textbf{Codec (s)} & \textbf{Network (s)} \\
    \midrule
    Synthetic-NeRF & 1.75  & 1.08  & 0.67 \\
    Mip-NeRF360   & 14.20 & 8.78  & 5.42 \\
    BungeeNeRF    & 25.98 & 16.07 & 9.91 \\
    \bottomrule
\end{tabular}
\end{table}

Table~\ref{tab:extra_analyses} extends the ablation study on Mip-NeRF360 (hr): (a) a nonlinear projection outperforms a single linear layer; (b) performance remains stable across a $4\times$ range of context dimensions $D^c$, and halving the default dimension increases size by only 2.6\%; (c) residual coding reduces both the anchor-feature alphabet and its coding cost; and (d) the standard deviations across three seeds are 0.02\,dB and 0.10\,MB for PSNR and size, respectively. Table~\ref{tab:dec_scaling} shows that decoding time grows with scene size. The recurrence is sequential only across the $N^C=5$ chunks within each anchor, while chunk $n$ is decoded for all anchors in a single batched step. Further speedups are possible through improvements to the arithmetic codec, such as chunk-parallel entropy coding.

\section{Limitations}
While RacerGS achieves substantial compression gains and high rendering fidelity, it has several limitations. The FiLM-modulated autoregressive GRU context model decodes anchor feature groups sequentially, which limits its applicability to real-time streaming scenarios. Additionally, like other anchor-based 3DGS methods, RacerGS is optimized per scene, so generalization to unseen scenes still requires retraining. Addressing these limitations through parallelizable context modeling and cross-scene generalization is a promising direction for future work.

\section{Societal impact}
\noindent\textbf{Positive impacts.} RacerGS substantially reduces the storage and transmission costs of high-fidelity 3D scene representations, which can lower the bandwidth and energy footprint of immersive applications such as virtual and augmented reality, telepresence, digital cultural heritage preservation, robotics simulation, and remote collaboration. By making compact 3D content more accessible on edge and mobile devices, our method can help democratize high-quality 3D media for users with limited compute or network resources.

\noindent\textbf{Negative impacts.} As a general-purpose 3D scene compression framework, RacerGS could be misused to more efficiently store, distribute, or stream 3D reconstructions of private spaces or individuals captured without consent, raising privacy and surveillance concerns. The lower bitrate also reduces the cost of disseminating photorealistic 3D content that may be repurposed for misleading or deceptive media. We encourage downstream practitioners to follow established consent, watermarking, and provenance practices, and to combine RacerGS with appropriate access control when deploying it on sensitive content.

\begin{table}[!t]
    \centering
    \small
    \caption{Encoding time for different components on the Mip-NeRF360 dataset.}
    \label{tab:encoding_time}
    \renewcommand{\arraystretch}{1.30}
    \resizebox{\columnwidth}{!}{%
    \begin{tabular}{ccccccccc}
    \toprule[1.5pt]
    $\lambda_r$ & \textbf{Total Time (s)} & $\bm{x}^a$ & $\bm{f}^a$ & $\bm{l}$ & $\bm{o}$ & $\mathcal{H}$ & $\bm{m}$ & Others \\
    \midrule
    $4e\text{-}3$ & 4.24 & 0.72 (17\%) & 1.93 (46\%) & 0.93 (22\%) & 0.45 (11\%) & 0.01 (0\%) & 0.01 (0\%) & 0.19 (4\%) \\
    $3e\text{-}3$ & 5.36 & 0.83 (15\%) & 2.59 (48\%) & 1.18 (22\%) & 0.56 (10\%) & 0.01 (0\%) & 0.01 (0\%) & 0.18 (3\%) \\
    $2e\text{-}3$ & 6.57 & 0.94 (14\%) & 3.25 (49\%) & 1.39 (21\%) & 0.69 (11\%) & 0.01 (0\%) & 0.01 (0\%) & 0.28 (4\%) \\
    $1e\text{-}3$ & 7.61 & 1.01 (13\%) & 3.88 (51\%) & 1.55 (20\%) & 0.82 (11\%) & 0.01 (0\%) & 0.01 (0\%) & 0.33 (4\%) \\
    $0.5e\text{-}3$ & 8.59 & 1.07 (12\%) & 4.40 (51\%) & 1.74 (20\%) & 0.96 (11\%) & 0.01 (0\%) & 0.01 (0\%) & 0.40 (5\%) \\
    \bottomrule[1.5pt]
    \end{tabular}}
    \end{table}

    \begin{table}[!t]
    \centering
    \small
    \caption{Decoding time for different components on the Mip-NeRF360 dataset.}
    \label{tab:decoding_time}
    \renewcommand{\arraystretch}{1.30}
    \resizebox{\columnwidth}{!}{%
    \begin{tabular}{ccccccccc}
    \toprule[1.5pt]
    $\lambda_r$ & \textbf{Total Time (s)} & $\bm{x}^a$ & $\bm{f}^a$ & $\bm{l}$ & $\bm{o}$ & $\mathcal{H}$ & $\bm{m}$ & Others \\
    \midrule
    $4e\text{-}3$ & 7.40 & 0.28 (4\%) & 3.69 (50\%) & 2.30 (31\%) & 0.94 (13\%) & 0.01 (0\%) & 0.01 (0\%) & 0.17 (2\%) \\
    $3e\text{-}3$ & 9.15 & 0.31 (3\%) & 4.71 (51\%) & 2.79 (30\%) & 1.21 (13\%) & 0.01 (0\%) & 0.01 (0\%) & 0.11 (1\%) \\
    $2e\text{-}3$ & 10.98 & 0.34 (3\%) & 5.72 (52\%) & 3.21 (29\%) & 1.54 (14\%) & 0.01 (0\%) & 0.01 (0\%) & 0.15 (1\%) \\
    $1e\text{-}3$ & 12.63 & 0.37 (3\%) & 6.65 (53\%) & 3.54 (28\%) & 1.82 (14\%) & 0.01 (0\%) & 0.01 (0\%) & 0.23 (2\%) \\
    $0.5e\text{-}3$ & 14.20 & 0.40 (3\%) & 7.55 (53\%) & 3.86 (27\%) & 1.98 (14\%) & 0.01 (0\%) & 0.01 (0\%) & 0.39 (3\%) \\
    \bottomrule[1.5pt]
    \end{tabular}}
    \end{table}

\begin{table}[!t]
    \centering
    \small
    \caption{Storage size (MB) of different components on the Mip-NeRF360 dataset.}
    \label{tab:total_param_size_supp}
    \setlength{\tabcolsep}{3.5pt}
    \begin{tabular}{ccccccccc}
    \toprule[1.5pt]
    $\lambda_r$ & \textbf{Total Size} & $\bm{x}^a$ & $\bm{f}^a$ & $\bm{l}$ & $\bm{o}$ & $\mathcal{H}$ & $\bm{m}$ & MLP \\
    \midrule
    $4e\text{-}3$ & 3.85 & 0.19 & 1.89 & 0.80 & 0.48 & 0.05 & 0.10 & 0.34 \\
    $3e\text{-}3$ & 4.92 & 0.20 & 2.57 & 1.01 & 0.63 & 0.06 & 0.11 & 0.34 \\
    $2e\text{-}3$ & 6.88 & 0.26 & 3.81 & 1.37 & 0.88 & 0.08 & 0.14 & 0.34 \\
    $1e\text{-}3$ & 10.24 & 0.35 & 5.94 & 1.93 & 1.35 & 0.10 & 0.23 & 0.34 \\
    $0.5e\text{-}3$ & 13.61 & 0.44 & 8.15 & 2.44 & 1.83 & 0.14 & 0.27 & 0.34 \\
    \bottomrule[1.5pt]
    \end{tabular}
    \end{table}

\begin{table}[!t]
    \caption{RacerGS results on the Synthetic-NeRF dataset for different $\lambda_r$ values.}
    \centering
    \scriptsize
    \begin{tabular}{cccccc}
    \toprule[1.5pt]
    $\lambda_r$ & Scenes & PSNR $\uparrow$ & SSIM $\uparrow$ & LPIPS $\downarrow$ & Size (MB) $\downarrow$ \\
    \midrule
    \multirow{8}{*}{$4e\text{-}3$} & chair & 33.54 & 0.960 & 0.025 & 0.28 \\
     & drums & 26.75 & 0.950 & 0.032 & 0.31 \\
     & ficus & 34.36 & 0.982 & 0.018 & 0.20 \\
     & hotdog & 35.65 & 0.971 & 0.033 & 0.24 \\
     & lego & 32.76 & 0.966 & 0.029 & 0.36 \\
     & materials & 30.21 & 0.950 & 0.025 & 0.31 \\
     & mic & 34.73 & 0.992 & 0.015 & 0.24 \\
     & ship & 32.90 & 0.937 & 0.147 & 0.27 \\
    \midrule
    \multirow{8}{*}{$3e\text{-}3$} & chair & 34.10 & 0.966 & 0.021 & 0.38 \\
     & drums & 26.59 & 0.950 & 0.033 & 0.42 \\
     & ficus & 34.66 & 0.983 & 0.017 & 0.32 \\
     & hotdog & 35.90 & 0.974 & 0.029 & 0.37 \\
     & lego & 33.26 & 0.969 & 0.026 & 0.48 \\
     & materials & 30.61 & 0.953 & 0.024 & 0.42 \\
     & mic & 35.10 & 0.992 & 0.013 & 0.33 \\
     & ship & 33.14 & 0.940 & 0.141 & 0.40 \\
    \midrule
    \multirow{8}{*}{$2e\text{-}3$} & chair & 34.47 & 0.971 & 0.019 & 0.54 \\
     & drums & 26.48 & 0.949 & 0.033 & 0.60 \\
     & ficus & 34.78 & 0.984 & 0.015 & 0.49 \\
     & hotdog & 36.21 & 0.977 & 0.027 & 0.52 \\
     & lego & 33.76 & 0.971 & 0.024 & 0.65 \\
     & materials & 30.92 & 0.955 & 0.022 & 0.60 \\
     & mic & 35.43 & 0.991 & 0.012 & 0.47 \\
     & ship & 33.39 & 0.946 & 0.136 & 0.53 \\
    \midrule
    \multirow{8}{*}{$1e\text{-}3$} & chair & 34.83 & 0.974 & 0.017 & 0.93 \\
     & drums & 26.32 & 0.946 & 0.034 & 1.02 \\
     & ficus & 34.82 & 0.981 & 0.015 & 0.88 \\
     & hotdog & 36.78 & 0.982 & 0.026 & 0.91 \\
     & lego & 34.42 & 0.971 & 0.022 & 1.01 \\
     & materials & 31.51 & 0.954 & 0.021 & 1.04 \\
     & mic & 35.72 & 0.985 & 0.016 & 0.56 \\
     & ship & 33.76 & 0.959 & 0.129 & 1.08 \\
    \midrule
    \multirow{8}{*}{$0.5e\text{-}3$} & chair & 34.83 & 0.981 & 0.025 & 1.40 \\
     & drums & 26.32 & 0.950 & 0.035 & 1.48 \\
     & ficus & 34.82 & 0.984 & 0.015 & 1.38 \\
     & hotdog & 36.78 & 0.993 & 0.025 & 1.33 \\
     & lego & 35.59 & 0.955 & 0.021 & 1.31 \\
     & materials & 32.42 & 0.943 & 0.024 & 1.48 \\
     & mic & 35.72 & 0.980 & 0.017 & 0.61 \\
     & ship & 33.76 & 0.973 & 0.121 & 2.54 \\
    \midrule
    $4e\text{-}3$ & \textbf{Avg.} & 32.61 & 0.964 & 0.041 & 0.28 \\
    $3e\text{-}3$ & \textbf{Avg.} & 32.92 & 0.966 & 0.038 & 0.39 \\
    $2e\text{-}3$ & \textbf{Avg.} & 33.18 & 0.968 & 0.036 & 0.55 \\
    $1e\text{-}3$ & \textbf{Avg.} & 33.52 & 0.969 & 0.035 & 0.93 \\
    $0.5e\text{-}3$ & \textbf{Avg.} & 33.78 & 0.970 & 0.035 & 1.44 \\
    \bottomrule[1.5pt]
    \end{tabular}
    \label{tab:per_scene_syncnerf}
    \end{table}

    \begin{table}[!t]
    \caption{RacerGS results on the Mip-NeRF360 dataset for different $\lambda_r$ values.}
    \centering
    \scriptsize
    \begin{tabular}{cccccc}
    \toprule[1.5pt]
    $\lambda_r$ & Scenes & PSNR $\uparrow$ & SSIM $\uparrow$ & LPIPS $\downarrow$ & Size (MB) $\downarrow$ \\
    \midrule
    \multirow{9}{*}{$4e\text{-}3$} & bicycle & 24.19 & 0.743 & 0.292 & 5.42 \\
     & bonsai & 32.00 & 0.927 & 0.188 & 2.87 \\
     & counter & 29.14 & 0.884 & 0.261 & 2.50 \\
     & flowers & 24.02 & 0.807 & 0.213 & 4.09 \\
     & garden & 26.20 & 0.759 & 0.295 & 5.14 \\
     & kitchen & 31.51 & 0.905 & 0.157 & 2.73 \\
     & room & 31.71 & 0.901 & 0.195 & 1.61 \\
     & stump & 25.61 & 0.694 & 0.396 & 5.43 \\
     & treehill & 23.17 & 0.586 & 0.362 & 4.86 \\
    \midrule
    \multirow{9}{*}{$3e\text{-}3$} & bicycle & 24.39 & 0.746 & 0.286 & 6.90 \\
     & bonsai & 32.08 & 0.929 & 0.186 & 3.48 \\
     & counter & 29.20 & 0.888 & 0.252 & 3.11 \\
     & flowers & 24.34 & 0.809 & 0.209 & 5.18 \\
     & garden & 26.29 & 0.763 & 0.289 & 6.58 \\
     & kitchen & 31.54 & 0.907 & 0.153 & 3.36 \\
     & room & 31.85 & 0.905 & 0.189 & 1.95 \\
     & stump & 25.60 & 0.698 & 0.382 & 7.00 \\
     & treehill & 23.20 & 0.590 & 0.356 & 6.72 \\
    \midrule
    \multirow{9}{*}{$2e\text{-}3$} & bicycle & 24.58 & 0.748 & 0.278 & 9.74 \\
     & bonsai & 32.14 & 0.930 & 0.186 & 4.96 \\
     & counter & 29.32 & 0.892 & 0.244 & 4.50 \\
     & flowers & 24.72 & 0.810 & 0.205 & 7.14 \\
     & garden & 26.37 & 0.769 & 0.276 & 9.27 \\
     & kitchen & 31.51 & 0.908 & 0.149 & 4.74 \\
     & room & 31.98 & 0.908 & 0.184 & 2.81 \\
     & stump & 25.52 & 0.702 & 0.372 & 9.84 \\
     & treehill & 23.16 & 0.596 & 0.347 & 8.92 \\
    \midrule
    \multirow{9}{*}{$1e\text{-}3$} & bicycle & 24.72 & 0.748 & 0.272 & 16.60 \\
     & bonsai & 32.17 & 0.929 & 0.193 & 9.82 \\
     & counter & 29.43 & 0.895 & 0.236 & 7.85 \\
     & flowers & 25.20 & 0.810 & 0.202 & 9.60 \\
     & garden & 26.38 & 0.782 & 0.259 & 12.63 \\
     & kitchen & 31.43 & 0.905 & 0.150 & 7.08 \\
     & room & 32.23 & 0.911 & 0.183 & 5.32 \\
     & stump & 25.40 & 0.702 & 0.366 & 11.89 \\
     & treehill & 23.06 & 0.599 & 0.344 & 11.37 \\
    \midrule
    \multirow{9}{*}{$0.5e\text{-}3$} & bicycle & 24.72 & 0.754 & 0.257 & 22.33 \\
     & bonsai & 32.17 & 0.903 & 0.193 & 19.86 \\
     & counter & 29.43 & 0.905 & 0.212 & 15.42 \\
     & flowers & 25.31 & 0.798 & 0.266 & 10.23 \\
     & garden & 26.78 & 0.800 & 0.237 & 14.17 \\
     & kitchen & 31.43 & 0.910 & 0.140 & 8.96 \\
     & room & 32.26 & 0.917 & 0.206 & 6.34 \\
     & stump & 25.40 & 0.709 & 0.350 & 12.65 \\
     & treehill & 23.06 & 0.605 & 0.328 & 12.57 \\
    \midrule
    $4e\text{-}3$ & \textbf{Avg.} & 27.51 & 0.801 & 0.262 & 3.85 \\
    $3e\text{-}3$ & \textbf{Avg.} & 27.61 & 0.804 & 0.256 & 4.92 \\
    $2e\text{-}3$ & \textbf{Avg.} & 27.70 & 0.807 & 0.249 & 6.88 \\
    $1e\text{-}3$ & \textbf{Avg.} & 27.78 & 0.809 & 0.245 & 10.24 \\
    $0.5e\text{-}3$ & \textbf{Avg.} & 27.84 & 0.811 & 0.243 & 13.61 \\
    \bottomrule[1.5pt]
    \end{tabular}
    \label{tab:per_scene_mipnerf}
    \end{table}

    \begin{table}[!t]
    \caption{RacerGS results on the Tanks\&Temples dataset for different $\lambda_r$ values.}
    \centering
    \scriptsize
    \begin{tabular}{cccccc}
    \toprule[1.5pt]
    $\lambda_r$ & Scenes & PSNR $\uparrow$ & SSIM $\uparrow$ & LPIPS $\downarrow$ & Size (MB) $\downarrow$ \\
    \midrule
    \multirow{2}{*}{$4e\text{-}3$} & truck & 25.82 & 0.861 & 0.178 & 4.46 \\
     & train & 23.08 & 0.833 & 0.212 & 3.92 \\
    \midrule
    \multirow{2}{*}{$3e\text{-}3$} & truck & 25.88 & 0.868 & 0.171 & 5.28 \\
     & train & 23.08 & 0.836 & 0.208 & 4.62 \\
    \midrule
    \multirow{2}{*}{$2e\text{-}3$} & truck & 25.94 & 0.876 & 0.166 & 6.71 \\
     & train & 23.08 & 0.838 & 0.205 & 5.91 \\
    \midrule
    \multirow{2}{*}{$1e\text{-}3$} & truck & 25.92 & 0.885 & 0.158 & 8.90 \\
     & train & 23.08 & 0.825 & 0.208 & 7.94 \\
    \midrule
    $4e\text{-}3$ & \textbf{Avg.} & 24.45 & 0.847 & 0.195 & 4.19 \\
    $3e\text{-}3$ & \textbf{Avg.} & 24.48 & 0.852 & 0.190 & 4.95 \\
    $2e\text{-}3$ & \textbf{Avg.} & 24.51 & 0.857 & 0.186 & 6.31 \\
    $1e\text{-}3$ & \textbf{Avg.} & 24.50 & 0.855 & 0.183 & 8.42 \\
    \bottomrule[1.5pt]
    \end{tabular}
    \label{tab:per_scene_tandt}
    \end{table}

    \begin{table}[!t]
    \caption{RacerGS results on the DeepBlending dataset for different $\lambda_r$ values.}
    \centering
    \scriptsize
    \begin{tabular}{cccccc}
    \toprule[1.5pt]
    $\lambda_r$ & Scenes & PSNR $\uparrow$ & SSIM $\uparrow$ & LPIPS $\downarrow$ & Size (MB) $\downarrow$ \\
    \midrule
    \multirow{2}{*}{$4e\text{-}3$} & playroom & 30.02 & 0.904 & 0.268 & 1.52 \\
     & drjohnson & 30.13 & 0.907 & 0.274 & 1.97 \\
    \midrule
    \multirow{2}{*}{$3e\text{-}3$} & playroom & 30.12 & 0.906 & 0.261 & 2.02 \\
     & drjohnson & 30.28 & 0.910 & 0.265 & 2.68 \\
    \midrule
    \multirow{2}{*}{$2e\text{-}3$} & playroom & 30.22 & 0.908 & 0.258 & 2.64 \\
     & drjohnson & 30.32 & 0.910 & 0.260 & 3.52 \\
    \midrule
    \multirow{2}{*}{$1e\text{-}3$} & playroom & 30.53 & 0.913 & 0.256 & 3.81 \\
     & drjohnson & 30.18 & 0.912 & 0.255 & 5.14 \\
    \midrule
    $4e\text{-}3$ & \textbf{Avg.} & 30.08 & 0.906 & 0.271 & 1.75 \\
    $3e\text{-}3$ & \textbf{Avg.} & 30.20 & 0.908 & 0.263 & 2.35 \\
    $2e\text{-}3$ & \textbf{Avg.} & 30.27 & 0.909 & 0.259 & 3.08 \\
    $1e\text{-}3$ & \textbf{Avg.} & 30.36 & 0.913 & 0.256 & 4.48 \\
    \bottomrule[1.5pt]
    \end{tabular}
    \label{tab:per_scene_deepblending}
    \end{table}

    \begin{table}[!t]
    \caption{RacerGS results on the BungeeNeRF dataset for different $\lambda_r$ values.}
    \centering
    \scriptsize
    \begin{tabular}{cccccc}
    \toprule[1.5pt]
    $\lambda_r$ & Scenes & PSNR $\uparrow$ & SSIM $\uparrow$ & LPIPS $\downarrow$ & Size (MB) $\downarrow$ \\
    \midrule
    \multirow{6}{*}{$4e\text{-}3$} & amsterdam & 26.75 & 0.854 & 0.239 & 8.23 \\
     & bilbao & 26.32 & 0.878 & 0.251 & 7.32 \\
     & hollywood & 26.52 & 0.875 & 0.358 & 8.03 \\
     & pompidou & 26.73 & 0.822 & 0.255 & 6.27 \\
     & quebec & 27.49 & 0.809 & 0.175 & 6.09 \\
     & rome & 25.25 & 0.831 & 0.252 & 7.17 \\
    \midrule
    \multirow{6}{*}{$3e\text{-}3$} & amsterdam & 27.01 & 0.870 & 0.215 & 10.42 \\
     & bilbao & 26.78 & 0.888 & 0.228 & 8.95 \\
     & hollywood & 26.38 & 0.876 & 0.338 & 9.21 \\
     & pompidou & 26.84 & 0.840 & 0.241 & 8.05 \\
     & quebec & 27.86 & 0.822 & 0.162 & 8.01 \\
     & rome & 25.55 & 0.846 & 0.233 & 11.64 \\
    \midrule
    \multirow{6}{*}{$2e\text{-}3$} & amsterdam & 27.28 & 0.872 & 0.203 & 13.82 \\
     & bilbao & 26.97 & 0.886 & 0.217 & 11.35 \\
     & hollywood & 26.42 & 0.874 & 0.322 & 12.17 \\
     & pompidou & 26.91 & 0.848 & 0.235 & 11.58 \\
     & quebec & 28.36 & 0.826 & 0.156 & 10.78 \\
     & rome & 25.94 & 0.857 & 0.223 & 16.32 \\
    \midrule
    \multirow{6}{*}{$1e\text{-}3$} & amsterdam & 27.28 & 0.876 & 0.179 & 17.45 \\
     & bilbao & 27.03 & 0.890 & 0.182 & 14.23 \\
     & hollywood & 26.42 & 0.767 & 0.313 & 15.15 \\
     & pompidou & 26.91 & 0.864 & 0.256 & 17.33 \\
     & quebec & 28.75 & 0.967 & 0.152 & 15.04 \\
     & rome & 26.69 & 0.840 & 0.212 & 17.71 \\
    \midrule
    $4e\text{-}3$ & \textbf{Avg.} & 26.51 & 0.845 & 0.255 & 7.19 \\
    $3e\text{-}3$ & \textbf{Avg.} & 26.74 & 0.857 & 0.236 & 9.38 \\
    $2e\text{-}3$ & \textbf{Avg.} & 26.98 & 0.861 & 0.226 & 12.67 \\
    $1e\text{-}3$ & \textbf{Avg.} & 27.18 & 0.867 & 0.216 & 16.15 \\
    \bottomrule[1.5pt]
    \end{tabular}
    \label{tab:per_scene_bungeenerf}
    \end{table}

%% file: tables/extra.tex
\begin{table}[!t]
\centering
\small
\caption{\textbf{Design analyses} on Mip-NeRF360.
(a)~Entropy-context projection.
(b)~Sensitivity to context dimension $D^c$.
(c)~Anchor-feature symbol statistics under attribute-lattice and residual coding.
(d)~Variability across three seeds (mean $\pm$ std).}
\label{tab:extra_analyses}
\footnotesize
\setlength{\tabcolsep}{7pt}
\renewcommand{\arraystretch}{1.08}

\begin{minipage}[t]{0.48\columnwidth}
\vspace{0pt}
\centering
\textbf{(a) Projection}\par\smallskip
\begin{tabular}{lcc}
\toprule
 & PSNR (dB) & Size (MB) \\
\midrule
None (raw $\bm{f}^h$) & 27.62 & 15.21 \\
Linear & 27.76 & 14.28 \\
\textbf{MLP (ours)} & \textbf{27.84} & \textbf{13.61} \\
\bottomrule
\end{tabular}
\end{minipage}
\hfill
\begin{minipage}[t]{0.48\columnwidth}
\vspace{0pt}
\centering
\textbf{(b) Context dimension}\par\smallskip
\begin{tabular}{ccc}
\toprule
$D^c$ & PSNR (dB) & Size (MB) \\
\midrule
12 & 27.80 & 13.96 \\
\textbf{24} & \textbf{27.84} & \textbf{13.61} \\
48 & 27.82 & 13.78 \\
\bottomrule
\end{tabular}
\end{minipage}

\par\medskip

\begin{minipage}[t]{0.48\columnwidth}
\vspace{0pt}
\centering
\textbf{(c) Anchor-feature symbols}\par\smallskip
\begin{tabular}{lcc}
\toprule
 & Attribute & \textbf{Residual} \\
\midrule
Symbol range & 1{,}023 & \textbf{287} \\
Empirical entropy (bits) & 8.27 & \textbf{5.22} \\
Coded bits / element & 8.44 & \textbf{5.38} \\
\bottomrule
\end{tabular}
\end{minipage}
\hfill
\begin{minipage}[t]{0.48\columnwidth}
\vspace{0pt}
\centering
\textbf{(d) Three seeds}\par\smallskip
\begin{tabular}{lcc}
\toprule
Method & PSNR (dB) & Size (MB) \\
\midrule
RacerGS & $27.84\pm0.02$ & $13.63\pm0.10$ \\
\bottomrule
\end{tabular}
\end{minipage}
\end{table}